\documentclass{new_tlp}
\usepackage{mathptmx}
\usepackage{url}
\usepackage{graphicx}
\usepackage{algorithmic}
\usepackage[noend,ruled,linesnumbered]{algorithm2e}
\usepackage{amsmath}
\usepackage{amssymb}
\usepackage{enumitem}
\usepackage{xcolor}
\usepackage{fancyvrb}
\let\proof\relax

\usepackage{amsthm}
\newcommand\bcmdtab{\noindent\bgroup\tabcolsep=0pt%
  \begin{tabular}{@{}p{10pc}@{}p{20pc}@{}}}
\newcommand\ecmdtab{\end{tabular}\egroup}

\newcommand{\FNOnt}{{\emph{FrameOnt}}\xspace}

  \title[Querying Knowledge via Multi-Hop English Questions]
        {Querying Knowledge via Multi-Hop English Questions}

  \author[Tiantian Gao, Paul Fodor, Michael Kifer]
         {Tiantian Gao, Paul Fodor, Michael Kifer\\
         Stony Brook University, Stony Brook, NY, USA\\
         \email{\{tiagao,pfodor,kifer\}@cs.stonybrook.edu}}

\newtheorem{definition}{Definition}
\newtheorem{example}{Example}
\begin{document}

\label{firstpage}

\maketitle

\begin{abstract}
  The inherent difficulty of knowledge specification and the lack of
  trained specialists are some of the key 
  obstacles on the way to making intelligent systems based on the knowledge
  representation and reasoning (KRR) paradigm commonplace.
  \emph{Knowledge and query authoring} using natural language, especially
  \emph{controlled} natural language (CNL), is one of the promising 
  approaches that could enable domain experts, who are not trained
  logicians, to both create formal knowledge and query it.
  In previous work, we introduced the \emph{KALM} system (Knowledge
  Authoring Logic Machine) that supports knowledge authoring (and simple
  querying)
  with very high accuracy that at present
  is unachievable via machine learning approaches.
  The present paper expands on the question answering aspect of KALM and
  introduces \emph{KALM-QA} (KALM for Question Answering) that is capable
  of answering much more complex English questions.
  We show that KALM-QA achieves 100\% accuracy on an extensive suite of
  movie-related questions, called \emph{MetaQA,} which contains almost 29,000
  test questions and over 260,000 training questions. We contrast this
  with a published machine learning approach, which falls far short of this high
  mark. It is under consideration for acceptance in TPLP.
  \end{abstract}


\section{Introduction}\label{intro}

Much of the world knowledge can be described by logical facts and rules,
and processed by intelligent \emph{knowledge representation and
  reasoning} (KRR) systems.  
Alas, eliciting knowledge from human experts and codifying it
in logic is often too big an undertaking, and the learning curve
for domain experts is too steep.
One would hope that such knowledge could be extracted from text, but this
goal is still out of reach.
Despite the advances in text understanding and information extraction (e.g., 
\cite{schmitz2012,angeli2015,Gomez08})
this technology is still far from delivering the accuracy needed for
KRR, which is very sensitive to errors.

Meanwhile, \emph{controlled natural languages} (CNL)
\cite{kuhn2014} were proposed as a possible solution to the aforesaid
learning curve problem in the belief that a natural language (NL) could be
more accessible to domain experts. The idea was that CNL sentences would
be usable as knowledge because,
as subsets of NLs with 
restricted grammars and interpretation rules, their
meaning can be captured in logic precisely.
Some of the better-known CNLs include
Attempto Controlled English (ACE) \cite{fuchs2008},
Processable English (PENG) \cite{schwitter2010}, and SBVR \cite{sbvr}.
CNL's restrictive grammars could, at times, be problematic, but this is not
a show-stopper for knowledge \emph{authoring}, since the difference between
knowledge \emph{authoring} and general knowledge \emph{acquisition} is
quite significant: whereas knowledge acquisition is about enabling machines to
understand the NL that humans write, knowledge authoring is about enabling
humans to write in NL that machines could understand.

A more serious obstacle for using CNLs in authoring
is that, by design, CNLs do not assume any kind
of background knowledge. For example, in ACE, the sentences \emph{Zoe
  Saldana appears in Avatar} and \emph{Zoe Saldana is an actress of Avatar}
would have different logical representations
and a query like \emph{Avatar has which actresses} will yield no answers.
Background knowledge could be written as CNL sentences in the
form of bridge rules, but it is naive to expect that domain experts would
know how to do it, and the amount of such background knowledge is prohibitive.
One of the main goals of knowledge authoring, which we started to address
in our previous work on the \emph{knowledge authoring logic
  machine} (KALM) \cite{Gao18,GaoFK18}, was to make this process scalable
and feasible while still relying on CNLs.

The previous incarnations of KALM supported
knowledge authoring and simple question answering 
with very high accuracy compared to the state-of-the-art machine learning
approaches, such as 
SEMAFOR \cite{journals/coling/DasCMSS14}, SLING \cite{ringgaard2017sling},
and Stanford CoreNLP \cite{manning-corenlp-2014}.
KALM achieves this
effect with the help of a sophisticated semantic layer on top of ACE, which
includes a formal, FrameNet-inspired \cite{JohnsonEtAl:01a} linguistic
ontology, \FNOnt,
that formalizes FrameNet's frames,
plus the linguistic knowledge graph BabelNet \cite{NavigliPonzetto:12aij}.
These linguistic resources are
relied on by an \emph{incrementally-learnable} semantic parser
that maps semantically equivalent CNL sentences
into the same \FNOnt frames and gives them 
\emph{unique logical representation} (ULR).

\noindent
\textbf{Contributions.}
The main contribution of this paper is introduction of 
KALM-QA, an extension of KALM that supports \emph{multi-hop} question
answering compared to just 1-hop queries in \cite{GaoFK18}. 
A \emph{multi-hop question} is one whose answering involves joining
(in the database sense) multiple relations, some of which could be
self-joins. Thus, answering a 3-hop query requires joining three
relations, as in
\emph{``Who wrote films that share actors with the film Anastasia''} and
\emph{``Who are the screenwriters that the actors in their movies also appear
  in the movie The Backwoods.''}
We use the MetaQA dataset \cite{metaqa-bib} that contains close to 29,000
multi-hop test queries and show that KALM-QA achieves the accuracy
of 100\% compared to the much lower accuracy of a machine learning
approach \cite{ZhangDKSS18}.

Second, we give a detailed description of the KALM-QA frame-semantic
parser, which is generated incrementally by structure learning. These
details did not appear in previous publications.
Finally, we note that MetaQA sentences use more general syntax than
what ACE accepts. \ref{stanford_parser} describes harnessing
the power of the Stanford Parser \cite{manning-corenlp-2014} 
for paraphrasing the MetaQA sentences (over
260,000 in total) to make them conform to ACE.

For the remainder of this paper,
Sections \ref{kalm} and \ref{metaqa} give the background on KALM and the MetaQA dataset.
Section \ref{structure_learning} describes KALM's frame-semantic parser in detail and
Section \ref{semantics} explains how KALM-QA translates
MetaQA multi-hop questions into logic.
Section \ref{experiment} presents experimental results,
Section \ref{related_works} discusses related works,
and Section \ref{conclusion} concludes the paper.
\ref{stanford_parser} describes the use of the Stanford Parser for paraphrasing
MetaQA sentences to conform to the ACE grammar. \ref{bgrule_estimation}
argues that KALM's approach to mediating between semantically equivalent
sentences is much more scalable than the more traditional approach of
specifying the equivalences via background bridge rules used in CNLs, as in
Attempto or SBVR.


\section{The KALM system}\label{kalm}

\begin{figure}[tb] 
 \centering
 \includegraphics[width=13.5cm]{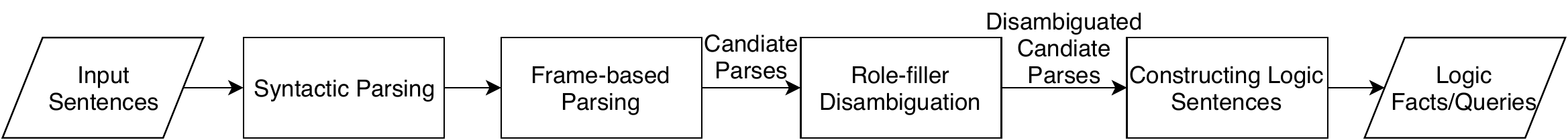}
 \caption{The KALM Architecture}
 \label{figure:kalm_architecture}
\end{figure}

KALM is a semantic framework that aims
to be a practically scalable solution for knowledge authoring.  KALM
users author knowledge using CNL sentences (ACE, to be specific) and KALM
ensures that semantically equivalent sentences
have the same logical representation.
This is achieved with the help of
linguistic and other information resources, which provide the
necessary background knowledge in a scalable fashion.
The architecture of KALM is shown in Figure \ref{figure:kalm_architecture}.

Not shown in the figure
is the new \emph{paraphrase} step, which
extends the syntax of Attempto\footnote{\url{http://attempto.ifi.uzh.ch/site/docs/syntax_report.html}} 
and lets KALM parse more complex MetaQA questions. This aspect is
discussed in \ref{stanford_parser}.

\noindent
\textbf{Syntactic parsing.} KALM uses Attempto Parsing Engine
(APE)\footnote{\url{https://github.com/Attempto/APE}} as the top level
parser. APE
extracts the syntactical information from sentences, including the
\emph{part of speech} of each word and the grammatical relations between
pairs of words.
This extracted information is represented by a list of stylized
first-order terms
known as a \emph{discourse representation structure}, or DRS.\footnote{\url{http://attempto.ifi.uzh.ch/site/pubs/papers/drs_report_66.pdf}}
\begin{example}\label{drs}
\textnormal{The sentence ``\emph{a director directs a film}'' has the following DRS}
\begin{verbatim}
    object(A,director,countable,na,eq,1)-1/2
    object(B,film,countable,na,eq,1)-1/5
    predicate(C,direct,A,B)-1/3
\end{verbatim}
\textnormal{
  Logically, this should be understood as a list of terms and the
  capitalized symbols as variables.
  The notation like \texttt{-1/2} says that the term refers to the second word
  in the first sentence. The
  \texttt{object}-term denotes an entity corresponding to a noun word
  and the \texttt{predicate}-term denotes an event/action
  corresponding to a verb.  The variables \texttt{A} and
  \texttt{B} denote the \emph{director}- and \emph{film}-entities, and the
  variable \texttt{C} denotes the \emph{directs}-event.
  When the same variable appears more than once, it serves as a
  cross-reference, as usual in logic.
  In a \texttt{predicate}-term, the 3rd and 4th arguments
  represent the \emph{subject} and \emph{object} of the corresponding
  event, respectively. 
  Arguments 4-6 in an \texttt{object}-term can be ignored for this paper.
  Thus,
  the above example shows that the \emph{directs}-event has
  the \emph{director}-entity as a subject and the \emph{film}-entity as an
  object.
}  \qed
\end{example}

\noindent
\textbf{Frame-based parsing.} Frame-based parsing takes DRS structures and
performs linguistic analysis trying to identify possible meanings of the
corresponding sentences---an idea inspired by FrameNet
\cite{JohnsonEtAl:01a}.
A \emph{frame}  \cite{fillmore01:FStextUnder} represents one or more
related semantic relationships
among entities, where each entity plays a particular \emph{role}.
While FrameNet is only an informal methodology where each frame is
described textually, KALM has a formal
ontology, \emph{FrameOnt}, which captures FrameNet's frames formally, as
logic formulas, and adds more frames as needed. 
\begin{example} \label{logical_frame}
\textnormal{
  The following \texttt{Movie} frame describes films and their attributes, including actors, directors, release years, and so on.
It is encoded by the following Prolog fact:}
\begin{verbatim}
    fp('Movie',[role('FilmNm',['bn:00034471n'],[]),
        role('Id',['bn:00045822n'],['Integer']),
        role('Actor',['bn:00001176n'],[]),
        role('Release Year',['bn:00078738n'],['Year']),
        role('Director',['bn:00027368n '],[]),
        role('Writer',['bn:00034485n'],[]),
        role('Genre',['bn:00037744n'],[])]).
\end{verbatim}
\textnormal{
  Much of this is self-explanatory except, possibly, the symbols of the
  form \texttt{bn:xxxxxxxxy}. These are BabelNet \emph{synsets}
  \cite{NavigliPonzetto:12aij}, which are similar to and derived from
  WordNet's synsets.
  Synsets are used because words typically have many meanings and the same
  meaning may be shared by many words. Thus, relying on words
  instead of synsets is likely to make information ambiguous and lead to
  wrong query answers.
  KALM is unique in the world of knowledge acquisition in its use of
  synsets.
  In addition, type constraints may be imposed on some roles. 
For instance, the \texttt{Release Year} role has the type constraint
\texttt{Year}, which insists that the role-filler word for this role
must be a valid year value.} \qed
\end{example}

A frame may represent several related relations, which
KALM captures via \emph{logical valence patterns} (\emph{lvps}).
An lvp consists of a \emph{lexical unit} and a set of \emph{grammatical patterns}, each associated with a 
frame role. A lexical unit is a word that could possibly ``trigger'' a
frame relation and thus indicate that the frame is possibly applicable to the
sentence. A grammatical pattern
represents a grammatical relation between the lexical unit and 
a frame role.
An lvp is applicable to a sentence if (1) the sentence contains the lexical unit and (2) the role-fillers for each frame role
can be extracted based on the respective grammatical patterns.
\begin{example}\label{lvp}
\textnormal{
  The following lvp says that the verb \emph{direct} is a lexical
unit of the frame \texttt{Movie}.
}
\begin{verbatim}
    lvp(direct,v,'Movie', [pattern('Director','verb->subject',required),
        pattern('FilmNm','verb->object',required)]).
\end{verbatim}
\textnormal{
The fillers for the roles \texttt{Director} and \texttt{FilmNm}
can be extracted by following the grammatical patterns
\texttt{'verb->subject'} and \texttt{'verb->object'}, respectively.
Each grammatical pattern acts as an independent extraction rule that works on the DRS structure.
For instance, the extraction rule \texttt{'verb->subject'}
(resp., \texttt{'verb->object'}) extracts the term
that represents the subject (resp., object) of the term that describes the lexical unit \texttt{direct}.
}
\textnormal{
  In the context of the DRS structure from Example \ref{drs}, 
the term corresponding to the lexical unit \texttt{direct} is \texttt{predicate(C,direct,A,B)}.
Applying the above lvp to that DRS
extracts a ``movie director'' relation where the word \emph{director} fills the
role \texttt{Director} and \emph{film} fills the role \texttt{FilmNm}.
}
\qed
\end{example}
We call the extracted relation a \emph{candidate parse}. 
Details on the construction of grammatical patterns and lvps will be discussed in Section \ref{structure_learning} .

\noindent
\textbf{Role-filler Disambiguation.}
Frame parsing produces \emph{candidate} parses by
identifying the lvps (i.e., semantic relationships) that a CNL
sentence represents. Alas, most of the triggered lvps might not make sense in
a particular situation. For instance, the sentence ``\emph{a director
  directs a film}'' triggers at least these two frames: making a film and
directing a corporation. In the first frame, ``\emph{film}'' is a filler for the
role \texttt{FilmNm} and in the second it fills the role \texttt{Corporation}.
The reason we know that this sentence is related to making a film and not
to directing a corporation is because a film is not a corporation.
Making such decisions is the job of role-filler disambiguation, which is
described in detail in \cite{Gao18} and relies on sophisticated algorithms
over the BabelNet knowledge graph \cite{NavigliPonzetto:12aij}.
Role-filler disambiguation also determines the right synset for each
role-filler. For instance, it will determine that the meaning of ``\emph{film}''
in our sentence is movie (\texttt{bn:00034471n}) 
and not photographic film (\texttt{bn:00034472n}).
The result of disambiguation consists
of \emph{disambiguated parses}.  
\begin{example}
\textnormal{
  For another example,
  consider the sentences (1) \emph{who makes a film} and (2) \emph{who makes a cake}.}
\textnormal{Both sentences share the same grammatical structure:
  \emph{make} is the main verb of the sentence, \emph{who} is the subject 
of the \emph{make}-event, \emph{film} and \emph{cake} are the objects of the \emph{make}-event for the first and second
sentences, respectively.
The first sentence implies the \texttt{Movie} frame while the second implies the \texttt{Cooking} frame.
However, since both sentences share the same grammatical structure,
the lvp  used to trigger the \texttt{Cooking} frame for the second sentence
also applies to the first.
This yields a wrong candidate parse  for sentence (1)
as an instance of the \texttt{Cooking}
frame, with \emph{who} as \texttt{Cook} and 
\emph{film} as \texttt{Food}. Since \emph{film} and \texttt{Food} are
semantically unrelated, role-filler disambiguation
will give a very low score to the second parse and it will get
pruned away.
} \qed
\end{example}
In the present paper, role-filler disambiguation is much simpler than in
\cite{Gao18}. First, the main terms in the MetaQA data set are used
unambiguously. Second, it includes an ontology that disambiguates
role-filler words with respect to the roles uniquely.
As a result, we do not need to use BabelNet here and
role-filler disambiguation here reduces to simply checking
class membership for the different entities.

\noindent
\textbf{Constructing logical forms.} This step maps the actual instances of
semantic relations extracted via lvps 
to their corresponding representations, called
\emph{unique logical representations (ULR)} for assertions of facts
and \emph{unique logical representation for queries (ULRQ)} for questions \cite{Gao18,GaoFK18}.
In this work, construction of ULRQ is much more complicated because 
multi-hop questions result in disambiguated parses consisting of multiple
lvps, which may even
come from different frames. In \cite{GaoFK18}, on the other hand,
we considered only 1-hop
queries, which correspond to single lvp disambiguated parses.
Details of multi-hop translation to ULRQ appear in Section \ref{semantics}.


\section{The MetaQA Dataset and Multi-Hop Questions}\label{metaqa}

The MetaQA dataset \cite{metaqa-bib} consists of a movie ontology derived from the WikiMovies
Dataset \cite{MillerFDKBW16} and three sets of question-answer pairs
written in natural language: 1-hop, 2-hop, and 3-hop queries.
In this paper, we focus on the last two as a more interesting challenge.
The movie ontology contains
information on films and their attributes including actors, directors,
writers, genres, release years, and so on.  MetaQA provides enough
information to differentiate between films that happen to have the same
name, but not namesake actors, writers, or directors.  Each film instance is
represented by a contiguous chunk of triples with the first argument
in a triple
being the film name, the second one of the
film's attributes (e.g.,  \emph{directed\_by}, \emph{has\_genre},
\emph{has\_imdb\_rating},
\emph{in\_language}, \emph{release\_year}, \emph{starred\_actors},
\emph{written\_by}), and the third argument being the attribute
value.  We assign a unique Id to each film instance and represent each triple as
a fact of the form \texttt{movie(AttrName1=AttrVal1,...)}.  For example, the MetaQA triple
\texttt{Kismet|\allowbreak directed\_by|William Dieterle} is represented as
\texttt{movie('FilmNm'='Kismet','Id'=72, 'Director'='William
  Dieterle')}, where \texttt{72} is the film Id.

\begin{example}
\textnormal{The sentence ``\emph{Who appears in a Steven Spielberg
    directed film}'' gets parsed into a conjunction of these two frame relations:
(1) \texttt{movie('FilmNm'=Title,'Id'=ID,'Director'\allowbreak ='Steven Spielberg')} and
(2) \texttt{movie('FilmNm'=Title,'Id'=ID,'Actor'=Y)}.
To construct the query to answer the above question we form a
  conjunction (a database join) of the two expressions. The result will be
  returned as a set of bindings for the variable \texttt{Y}. 
} \qed
\end{example}

\paragraph{Structure of MetaQA.}
The \emph{training set} of MetaQA has 118,980 2-hop and 114,196 3-hop
questions labeled with the expected query answers.  The \emph{testing set}
has 14,872 2-hop and 14,274 3-hop questions---also labeled with expected
answers.  The \emph{movie ontology}   itself contains information about 17,341
distinct movies including their directors, writers, actors, and other
properties, as described earlier.

\paragraph{Problems with MetaQA.}
As it turned out, a large fraction of both training and testing queries in
MetaQA are mislabeled, i.e., the answers provided are wrong. In the
training set, 13,522 out of 118,980 2-hop queries, or 11.36\%, are definitely
mislabeled, but we suspect this number is as high as 25,487 (21.42\%).
For 3-hop queries, the situation is even worse: 36,645 questions out of 114,196 
(32.09\%) are labeled with wrong answers, but this number is likely as high
as 64,089 (56.12\%).
In the testing set, the situation is similar:
3,178 2-hop queries out of 14,872 (21.37\%) are mislabeled and for 3-hop
queries this number is 8,062 out of 14,274 (56.48\%).

Most of the labeling errors are due to the fact that MetaQA's designers
\cite{ZhangDKSS18} failed to take into account that the movie ontology has
many movies with the same title. As a result, queries like ``\emph{Who
  co-acted (or co-directed) with X in a movie}'' are likely to have wrong
answers if X acted in movie M1, Y acted in M2, and M1 and M2 happen to have
exactly the same title.  There are, for example, three movies named ``Jane
Eyre,'' which trigger this kind of errors.  The existence of such errors is
easily checked automatically with KALM and this is where the aforesaid
counts of  certain errors in the training set (13,522 and 36,645) come
from.

In the testing set, the same problem
accounts for the bulk of the wrong expected answers: 1,707 and 4,705. The remaining
errors are due to more subtle reasons such as confusion between a certain
number being a movie title vs. it being a movie release year (e.g., the
movies ``\emph{1941}'', ``\emph{2010}''), failure to recognize that, say, \emph{Thomas Mann}
is a writer and not a director of ``\emph{Death in Venice}'' (which causes the
question ``\emph{What are the release years of the movies directed by Thomas
Mann?}'' to mistakenly return \emph{1971}), or the failure to realize that someone
could be both a writer and an actor in the same movie (thus causing the answer
``\emph{Bob Peterson}'' to be missing in MetaQA's answer for the query ``\emph{Who are
the actors in the films written by Bob Peterson?}'').
For these latter, more subtle, issues we checked manually about
50\% of the suspected errors (736 of 1,471 2-hop and 1628 of 3,357 3-hop)
for the testing set and verified that KALM provides correct answers while
MetaQA does not. The number of suspected errors in the training set is too
large to be checked manually, so we checked only a few dozen to confirm our
hypothesis. 

The large number of errors in the original MetaQA puts in question the
accuracy of the results reported in \cite{ZhangDKSS18} to which we return
in Section~\ref{experiment}.
A corrected version of MetaQA data set can be found in reference \cite{Gao:2019}.


\section{Constructing a KALM Semantic Parser via Structural Learning}\label{structure_learning}

This section shows how KALM's frame-semantic parser is constructed by
incremental structure-learning.
First, we show how grammatical patterns for role-filler extraction are
learned from training sentences and then 
how lvps are generated based on annotated training
sentences.

\subsection{Learning Grammatical Patterns for Role-filler Extraction}\label{grammatical_patterns}


Example \ref{lvp}
of Section~\ref{kalm}
illustrates how a
grammatical pattern like \texttt{verb->subject} is used to extract
the subject of the \emph{direct}-event.
Technically, it is done by these
Prolog extraction rules:
\begin{verbatim}
  extraction_rule('verb->subject',DRS,LexicUnitTerm,RoleFillerTerm):- 
      get_subject_from_verb(DRS,LexicUnitTerm,RoleFillerTerm). 
  get_subject_from_verb(DRS,VerbTerm,SubjectTerm):-
      get_arg3_from_verb_term(VerbTerm,Arg3),
      SubjectTerm = object(Arg3,_,_,_,_,_),
      member(SubjectTerm,DRS).
\end{verbatim}
The first argument in \texttt{extraction\_rule/4} 
is a string representing the grammatical pattern.
Given the DRS for a sentence and
the lexical unit term as the input,
the extraction rule outputs the extracted role-filler term 
by calling a \emph{utility predicate} \texttt{get\_subject\_from\_verb/3}
that: (1) extracts argument 3, \texttt{Arg3}, from
of the verb term \texttt{VerbTerm} (the verb term here describes the
lexical unit and \texttt{Arg3} is the subject of that verb); 
and (2) finds an \texttt{object}-term in DRS whose first argument,
\texttt{Arg1} (that represents the \texttt{object}-term),
matches \texttt{Arg3} (the subject of the \texttt{predicate}-term).
If the \texttt{DRS}  argument is as in Example \ref{drs} and the lvp is as
in Example \ref{lvp}, the lexical unit term in question would be
\texttt{predicate(C,direct,A,B)}, the \texttt{subject}-term would be
\texttt{A}    and
the matching \texttt{object}-term would be \texttt{object(A,director,count\allowbreak able,na,eq,1)}. 

Grammatical patterns and the corresponding extraction rules are learned through annotated training sentences. 
An annotated sentence is a CNL sentence with lexical information attached
to it, which includes the frame name, the lexical unit, and
frame roles. Conceptually an annotated
sentence looks like this:
{\color{purple}[Movie]}\emph{a director}{\color{blue} $_\text{Director}$} \emph{directs}{\color{red} $_\text{lexical\_unit}$}
\emph{a film}{\color{blue} $_\text{Film}$}.
Later, in Example~\ref{annotation}, we will see the actual
structure used to represent annotated sentences.

The structure-learning process starts with
ACE parsing the raw CNL sentence found in a provided annotated training
sentence, which yields a DRS.
Then, for each pair of terms (\emph{lexical-unit,role-filler}) in that DRS
structure,
KALM produces and saves the corresponding extraction rule, like the one at
the start of this subsection.
(Recall that the aforesaid lexical unit and the role fillers are marked in the training
sentence explicitly.)
Creation of extraction rules has three parts: (1) construction of the library
of utility rules, which is done in advance; (2) reachability reasoning in
DRS;
and (3) construction of the actual extraction rules.

\noindent
\textbf{Utility predicates and rules.} These rules form the core library of
the KALM parser; they are used for commonly occurring 
manipulations with DRS structures and do not depend on the application domain.
Each utility predicate is defined by a single
utility rule, which handles a concrete
navigation scenario among terms that share variables
within the same DRS.
\begin{example}\label{ex-utilpred}
\textnormal{A prepositional word is represented by the term \texttt{modifier\_pp(A,Lexem,B)} 
where the first argument, \texttt{A},
denotes the verb being modified, the second argument represents
the propositional word and its lexical information,
and the third argument denotes the object in the prepositional phrase.
Due to the linguistic
purpose of a \texttt{modifier\_pp} term, it can be reached via two DRS terms:
via the shared variable \texttt{A} in 
a term like
\texttt{predicate(A,Verb,SubjRef,ObjRef)},
and via the shared variable \texttt{B} in a term like
\texttt{object(B,Noun,Class,Unit,Op,Count)}. 
These two cases lead to two utility rules.
The first takes the first argument \texttt{Arg1} from a \texttt{predicate}-term
and then finds a \texttt{PrepositionalTerm} (of the form
\texttt{modifier\_pp(A,Lexem,B)})
in DRS whose first argument matches \texttt{Arg1}.}
\begin{verbatim}
  get_preposition_from_verb(DRS,VerbTerm,PrepositionalTerm):-
      get_arg1_from_verb_term(VerbTerm,Arg1),
      PrepositionalTerm = modifier_pp(Arg1,_,_),
      member(PrepositionalTerm,DRS).
\end{verbatim}
\textnormal{The second rule takes the third argument \texttt{Arg3} from
  \texttt{PrepositionalTerm}
and finds an \texttt{object}-term in DRS whose first argument matches \texttt{Arg3}.}
\begin{Verbatim}[commandchars=\\\{\}]
\emph{
  get_dependent_from_preposition(DRS,PrepositionalTerm,DependentTerm):-
}
\emph{
      get_arg3_from_prepositional_term(PrepositionalTerm,Arg3),
}
\emph{
      DependentTerm = object(Arg3,_,_,_,_,_),
}
\emph{
      member(DependentTerm,DRS).   \qed
}
\end{Verbatim}
\end{example}
\noindent
Each utility rule is associated with a unique grammatical pattern (written
as a string) via the predicate
\texttt{utility\_grammatical\_pattern(UtilityPredicate,GrammaticalPattern)}.
For instance, the above utility predicates
are associated with the grammatical patterns
\texttt{verb->pp} and \texttt{pp->dep}, respectively.
This is the source of the grammatical patterns that we saw earlier in
\texttt{extraction\_rule/4}. We will come back to this at the end
of this subsection.

\noindent
\textbf{Reachability reasoning in DRS}. To see how one term in a DRS
is connected to another via a sequence of intermediate
variables and terms, we reduce this problem to a graph reachability by embedding the DRS in a graph structure: 
\begin{enumerate}
\item For each term in DRS, create a node in the graph.
\item For any pair of nodes $n_{1}$, $n_{2}$ that represent the DRS subterms
  $p_{1}$, $p_{2}$ \emph{that share a variable},
  create a directed labeled edge $n_{1} \longrightarrow n_{2}$ and back.
  The label of the edge $n_{1} \longrightarrow n_{2}$ is the
  position of the shared variable in $p_{1}$ and
  the edge $n_{2} \longrightarrow n_{1}$ is labeled with the position of that
  variable in $p_{2}$.
\end{enumerate}

Let $n_{i}$ be the node for the term that represents a lexical
unit and $n_{j}$ be the node for a role-filler term.
Then, the problem of constructing the extraction rule for the role-filler
reduces to finding
the shortest path from $n_{i}$ to $n_{j}$.
The above graph is well-defined, \emph{if} no term in a DRS has
repeated variables, which is always the case for sentences that have no
co-occurrences of the same entities or anaphoras (e.g., ``\emph{Fido eats
  Fido}'' or ``\emph{Fido eats himself}'' would violate this restriction).
This restriction ensures that any pair of terms in DRS shares \emph{at most one}
variable.
It is needed \emph{only} for training sentences and 
in practice is easily achievable via a trivial paraphrase, since the
training set is under the control of the knowledge engineer. Once learning
is done, testing sentences \emph{can} have anaphoras and entity
co-occurrences with no restrictions.

\begin{figure}[tb]
 \centering
 \includegraphics[width=6.5cm]{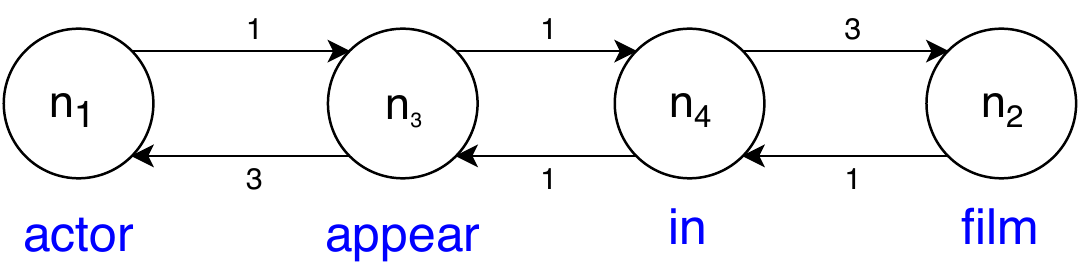}
 \caption{The DRS embedding for the sentence \emph{an actor appears in a film}.}
 \label{figure:drs_embedding}
\end{figure}

\begin{example}\label{drs:embedding}
\textnormal{The annotated sentence {\color{purple}[Movie]}\emph{an actor{\color{blue}
      $_\text{Actor}$} appears{\color{red} $_\text{lexical unit}$} in a
    film{\color{blue} $_\text{Film}$}} has the following DRS:}
\begin{verbatim}
    object(A,actor,countable,na,eq,1)-1/2
    object(B,film,countable,na,eq,1)-1/6
    predicate(C,appear,A)-1/3
    modifier_pp(C,in,B)-1/4
\end{verbatim}
\textnormal{Let $n_{1}$, $n_{2}$, $n_{3}$, and $n_{4}$ denote each of the
  terms in DRS, respectively. 
The DRS is represented by the graph shown in Figure \ref{figure:drs_embedding}. 
Learning the grammatical pattern that connects the lexical unit
\emph{appear} to \emph{film}
reduces to finding the shortest path from $n_{3}$ to $n_{2}$, i.e., $n_{3}$ $\xrightarrow[]{1}$ $n_{4}$ $\xrightarrow[]{3}$ $n_{2}$.} \qed
\end{example}

\noindent
\textbf{Construction of extraction rules}. 
Extraction rules are generated based on the shortest path from a lexical unit node to a role-filler node.
As is shown in Figure \ref{figure:drs_embedding}, two adjacent nodes plus one of their connected edges uniquely defines a scenario
where one term is connected to the other via a common variable in DRS. 
Thus, the body of the extraction rule is formed as a join of the
corresponding utility predicates for every pair of adjacent edges
in the path.
For instance, for the path
$n_{3}$ $\xrightarrow[]{1}$ $n_{4}$ $\xrightarrow[]{3}$ $n_{2}$ in 
Example \ref{drs:embedding}, KALM generates the following extraction rule:
\begin{verbatim}
extraction_rule('verb->pp,pp->dep',DRS,LexicUnitTerm,RoleFillerTerm):- 
    get_preposition_from_verb(DRS,LexicUnitTerm,PrepositionTerm),
    get_dependent_from_preposition(DRS,PrepositionTerm,RoleFillerTerm).
\end{verbatim}
The grammatical pattern in the first argument is a concatenation of the
grammatical patterns associated with the two utility
predicates in the body of the extraction rule, which were introduced right
after the end of Example~\ref{ex-utilpred}.

\subsection{Learning Lvps for Frame-semantic Parsing}
Lvps are learned from annotated training sentences. An annotated sentence
is a term that includes the original CNL sentence,
the name of a semantic frame that matches the sentence, the word index of
the lexical unit, and the word indices of each role-filler word and the
respective frame role.
An annotated sentence is represented by a Prolog fact.
Note: the same English sentence may carry several different bits of
information and thus may match several different frames. In this case,
there might be several annotated sentences for the same CNL sentence.
\begin{example}\label{annotation}
\textnormal{Consider the annotated training sentence:
\texttt{annotation('An actor appears in a film','Movie',3,[['Actor',2],['FilmNm',6]]).}}
\textnormal{It relates to the \texttt{Movie} frame with the
  word \emph{appears} serving as the lexical unit. The words
\emph{actor} and \emph{film} fill the roles \texttt{Actor} and
\texttt{Film}, and the numbers are word indices in the sentence.} \qed
\end{example}

KALM feeds annotated sentences to the \texttt{learn\_lvp} rule, which
generates extraction rules for each frame role in
the annotation and then forms an lvp. At a high level, the
\texttt{learn\_lvp} rule looks as shown below:
\begin{Verbatim}[commandchars=\\\{\}]
learn_lvp(annotation(Sentence,FrameName,LexicUnitIdx,RoleFillerIdxes)) :-
  ace_to_drs(Sentence,DRS),  %% \emph{invoke Attempto parser}
  grammatic_patterns(DRS,LexicUnitIdx,RoleFillerIdxes,GrammaticPatterns),
  construct_lvp(DRS,LexicUnitIdx,FrameName,GrammaticPatterns).
\end{Verbatim}
In the rule body, the predicate \texttt{ace\_to\_drs} calls the ACE parser
to produce the DRS of the input sentence.
Using the DRS and the word indices of the lexical unit and each of
the role-filler words,
the predicate \texttt{grammatic\_patterns} generates the grammatical patterns and the corresponding extraction rules
for each frame role. The grammatical patterns are passed to the next
subgoal in a variable, while the extraction rules are saved in KALM
parser's library.
Finally, the predicate \texttt{construct\_lvp} takes the lexical unit,
the frame name, and grammatical patterns
constructs the requisite lvp and asserts it in the database.
For instance, the annotated sentence of
Example \ref{annotation} will yield the following lvp:
\begin{verbatim}
    lvp(appear,v,'Movie', [pattern('Actor','verb->subject',required),
        pattern('Film','verb->pp,pp->dep',required)]).
\end{verbatim}
A similarly annotated sentence ``\emph{a director directs a film}''
yields an lvp that we met in Example~\ref{lvp}.

Once the necessary lvps are learned, KALM uses the following query
to parse new (annotation-free) sentences; it finds the lvps that match the
sentence and outputs a set of candidate parses:
\begin{verbatim}
    ?- parse_sentence(Sentence,CandidateParseList).
\end{verbatim}
The \texttt{parse\_sentence} rule can be described, at a high level, as
follows:
\begin{verbatim}
  parse_sentence(Sentence,CandidateFrameList) :-
      ace_to_drs(Sentence,DRS),
      get_candidate_lexical_units_from_drs(DRS,CandidateLexicUnitList),
      get_lvps(CandidateLexicUnitList,LvpList),
      apply_lvps_to_drs(DRS,LvpList,CandidateFrameList).
\end{verbatim}
The first step yields a DRS, as before.
Then, \texttt{get\_candidate\_lexical\_units\_from\_drs/2} finds all
\emph{candidate} lexical units in the DRS that can possibly trigger a frame.
Each such candidate is
represented by a tuple that includes a lexeme and a part-of-speech.
Third, \texttt{get\_lvps/2} finds all lvps whose lexical
unit matches
a candidate lexical units in \texttt{CandidateLexicUnitList}. 
Finally, \texttt{apply\_lvps\_to\_drs/3}
applies each lvp in \texttt{LvpList} to the DRS
to see which ones succeed. 
Section \ref{grammatical_patterns} explained how grammatical patterns are associated 
with rules for extracting role-filler words.
The predicate \texttt{apply\_lvps\_to\_drs/3} takes
each lvp in \texttt{LvpList} and applies the extraction rules matching
each \texttt{pattern}-term in the lvp to find candidate frames that fit the
input \texttt{Sentence}. 
An lvp may fail because its grammatical pattern is such
that an associated extraction rule fails to extract the expected filler
word or the extracted words have the wrong type or fail other constraints.
The successful lvps will yield frame relation instances, i.e.,
candidate parses.


\section{Capturing the Meaning of Multi-Hop Questions in Logic}\label{semantics}

This section shows how candidate parses of multi-hop questions are
translated into logical queries.

\begin{definition}
A \textbf{candidate parse} is a tuple of the form $(frame\_name,
lexical\_unit,\allowbreak \{(frame\_role_{i}, \allowbreak
role\_filler_{i},grammatic\_pattern_{i})\})$ ($1 \leq i \leq n$) where the
first element is a frame name, the second is the
word index of the lexical unit used to extract the frame relation, and the
third element is a set of tuples, each consisting of three elements:
a frame role, the word index of the role-filler word and the grammatical
pattern used to extract the role-filler. \qed
\end{definition}

English sentences for multi-hop questions typically result in multiple
candidate parses that form complex inter-relationships.
These relationships must be taken into account in order to translate these
sets of candidate parses into correct logic queries.

\begin{definition} 
Let $f\ =\ (name, lu, \{(r_{1},w_{1},p_{1}),(r_{2},w_{2},p_{2}),\ldots,\allowbreak (r_{m},w_{m},p_{m})\})$
and $f'\ =\ (name',\allowbreak lu', \{(r_{1}',w_{1}',p_{1}'),(r_{2}',w_{2}',p_{2}'),\ldots,\allowbreak (r_{n}',w_{n}',p_{n}')\})$
be two candidate parses where $m \leq n$. We say that $f'$ \textbf{subsumes} $f$ if $name = name'$ and 
$\{(r_{1},w_{1}),\ldots,\allowbreak (r_{m},w_{m})\} \subseteq \{(r_{1}',\allowbreak w_{1}'),\ldots,\allowbreak (r_{n}',w_{n}')\}$. \qed
\end{definition}
\noindent
Subsumed parses represent the relationships implied by subsuming
parses and can be ignored.

\begin{definition}
Let $f\ =\ (name, lu, \{(r_{1},w_{1},p_{1}),(r_{2},w_{2},p_{2}),\ldots,\allowbreak (r_{m},w_{m},p_{m})\})$
and $f'\ =\ (name',\allowbreak lu', \{(r_{1}',w_{1}',p_{1}'),(r_{2}',w_{2}',p_{2}'),\ldots,\allowbreak (r_{n}',w_{n}',p_{n}')\})$
be two candidate parses. We say that $f$ and $f'$ are \textbf{alternatives}
if:~ \emph{(1)}   $lu\ =\ lu'$ and 
$\{(w_{1},p_{1}),(w_{2},p_{2}),\ldots,\allowbreak (w_{m},p_{m})\} \subseteq
\{(w_{1},\allowbreak p_{1}),(w_{2},p_{2}),\ldots,\allowbreak (w_{n},p_{n})\}$
or vice versa; and \emph{(2)} $f$, $f'$ do not subsume each other.

\noindent
This implies that either $name \neq name'$ or some of the extracted words
fill different roles in the two parses. Thus, the two parses
represent different frame relations among the extracted words. \qed
\end{definition}

\begin{definition}
A set of candidate parses $F = \{f_{1}$, $f_{2}$, $\ldots$, $f_{n}\}$ is
a \textbf{set of pairwise alternatives} if for any $i$ and $j$, where $i \neq j$, 
$f_{i}$ and $f_{j}$ are alternatives. $F$ is a \textbf{maximal set
  of pairwise alternatives} if there is no $F'$ such that $F'$ is pairwise alternative 
and $F \subset F'$. \qed
\end{definition}

Let $F = \{f_{1}$, $f_{2}$, $\ldots$, $f_{n}\}$ be a set
of candidate parses for a CNL sentence $C$.
The \emph{unique logical representation for queries} 
(\emph{ULRQ}) for $C$ is obtained via the following steps:
\begin{itemize}
\item[1.] 
  Prune invalid candidate parses from $F$ via role-filler
  disambiguation. Let the result be $F'$.
\item[2.]
  Delete all
  $f_{i} \in F'$ that is subsumed by some $f_{j} \in F'$ ($i \neq j$),
  yielding $F''$.
\item[3.] Let $\overline{F}$ be $\{F_{1}$, $\ldots$, $F_{k}\}$, where
  each $F_{i} \subseteq F''$ ($1 \leq i \leq k$) is a maximal set of pairwise
  alternatives.
 Use the mapping $\phi$, described below, to map
 $\overline{F}$ to an ULRQ.
\end{itemize}
Conceptually, $\overline{F}$ represents a multi-hop question (if $k>1$),
where each $F_{i}$ is a single hop represented by a union of base
relations. This is reflected in the definition of the
mapping $\phi$ below:
\begin{itemize}[leftmargin=0cm]
\item[]
  $\phi(f) = $ {\emph{the relation associated with $f$}}   \hfill if $f$ is a single
  candidate parse \quad \quad \quad \quad\qquad \qquad  {~~~}
\item[] $\phi(\{f_{1}$, $\ldots$, $f_{k}\}) = \phi(f_{1}) \vee \ldots \vee
  \phi(f_{k})$ \hfill if $\{f_{1}, ..., f_{k}\}$ is a set of
  pairwise alternatives \quad\qquad ~
\item[] $\phi(\{F_{1}$, $\ldots$, $F_{m}\}) = \phi(F_{1}) \wedge
  \ldots \wedge \phi(F_{m})$  \hfill if each $F_{i}$ is a maximal
  set of pairwise alternatives~~~~~
\end{itemize}
If $f$ is a single candidate parse, $\phi$ maps it to a ULRQ as described
in \cite{GaoFK18}---a relation uniquely associated with
$f$'s lvp with arguments filled in by the extracted words (such as
\texttt{movie('FilmNm'='M','ID'=\_,director=D)}).
A set of pairwise alternatives is treated by $\phi$ as a single hop and
is mapped to a union of relations.
A set of sets of pairwise alternatives
is mapped by $\phi$ into a database join of the queries that correspond to the
constituent hops.
To illustrate,
we give two examples of translation of multi-hop questions to ULRQ.
\begin{example}
\textnormal{Consider the sentence ``\emph{Who wrote a film that shares a
    director with Titanic?}''. This sentence has four candidate parses:}
\begin{itemize}[leftmargin=.0cm]
\item[] $f_{1}$: \textnormal{\texttt{(Movie,2,[(FilmNm,4,'verb->subject'), (Writer,2,'verb->object')])}}
\item[] $f_{2}$: \textnormal{\texttt{(Movie,6 [(FilmNm,4,'verb->subject'),(Director,8,'verb->object')])}}
\item[] $f_{3}$: \textnormal{\texttt{(Movie,6,[(FilmNm,10,'verb->pp,pp->dep'),}}\textnormal{\texttt{(Director,8,'verb->object')])}}
\item[] $f_{4}$: \textnormal{\texttt{(Distinct,6,[(Item1,4,'verb->subject'),}}\textnormal{\texttt{(Item2,10,'verb->pp,pp->dep')])}}  
\end{itemize}
\textnormal{with the maximum pairwise alternative sets $F_{1} =
  \{f_{1}\}$, $F_{2} = \{f_{2}\}$, $F_{3} = \{f_{3}\}$, and
  $F_{4} = \{f_{4}\}$, yielding
  the following query:}
\begin{verbatim}
  ?- movie('FilmNm'=Title1,'Id'=ID1,'Writer'=V2),
     movie('FilmNm'=Title1,'Id'=ID1,'Director'=V3),
     movie('FilmNm'='Titanic','Id'=ID2,'Director'=V3),
     distinct('Item1'=ID1,'Item2'=ID2).
\end{verbatim}
\textnormal{where capitalized symbols are variables.
  Here \texttt{distinct} is a domain-independent built-in frame
  that indicates that the entities involved are distinct. It is triggered
  by the lexical units such as 
  \emph{...shared...with...}, \emph{...different...from...}, and the like. 
} \qed
\end{example}

\begin{example}
\textnormal{The question ``\emph{Who is an actor of Pascal Laugier?}''
  has these two candidate parses:}
\begin{itemize}[leftmargin=.8cm]
\item[] $f_{1}$: \textnormal{\texttt{(Coop,4,[(Actor,1,'object->verb,verb->subject'),
\item[] \qquad \qquad \qquad \, (Director,7,'lobject->rel,rel->robject')])}}
\item[] $f_{2}$: \textnormal{\texttt{(Coop,4,[(Actor,1,'object->verb,verb->subject'),
\item[] \qquad \qquad \qquad \, (Writer,7,'lobject->rel,rel->robject')])}}
\end{itemize}
\textnormal{The frame \texttt{Coop} here represents the concept of
  cooperation such as writers co-writing a movie, actors co-acting 
  or playing in a movie directed by somebody, etc.
  Since this is a composite concept, it must be defined via a background
  rule, which is done using the \emph{same CNL}.  
  For example, the background rule for the concept expressed by the first
  candidate parse is
  ``\emph{If an actor plays in a film that is directed by a director and
    the actor is different from the director
    then the actor is an actor of the director}''
  and KALM would translate this into the following Prolog rule:}
\begin{Verbatim}[commandchars=\\\{\}]
\emph{
    coop('Actor'=V1,'Director'=V2):-
}
\emph{
         movie('FilmNm'=Title1,'Id'=ID1,'Actor'=V1),
}
\emph{
         movie('FilmNm'=Title1,'Id'=ID1,'Director'=V2),
}
\emph{
         distinct('Item1'=V1,'Item2'=V2).
}
\end{Verbatim}
\textnormal{Since \emph{Pascal Langier} is both a writer and a director in the MetaQA
ontology, the two parses are alternatives to each other,
which correctly captures the intent that the query should return those
actors who play in movies that are either directed or written by
\emph{Pascal Laugier}.
This sentence has just
one maximum pairwise alternative set, $F_{1} = \{f_{1}, f_{2}\}$ and the
corresponding query is}
\begin{Verbatim}[commandchars=\\\{\}]
  \emph{
  ?- coop('Actor'=V1,'Writer'='Pascal Laugier');
  }
  \emph{
     coop('Actor'=V1,'Director'='Pascal Laugier'). \qed
   }
\end{Verbatim}
\end{example}


\section{Experiments}\label{experiment}
In this section, we first describe the evaluation of KALM-QA on the MetaQA dataset \cite{metaqa-bib}
and then show how we verified that queries generated by KALM-QA are 100\% correct.

\subsection{Evaluation of KALM-QA on the MetaQA Dataset}
We now discuss our experiments using the extensive MetaQA Vanilla dataset \cite{metaqa-bib}
of 2- and 3-hop questions about movies, described in
Section~\ref{metaqa}, and compare our results with
the machine learning approach reported in \cite{ZhangDKSS18}.

\paragraph{Settings.} 
Prior to the start of our experiments, KALM had only 50 frames with
217 lvps having nothing to do with movies. Therefore, we expanded the
semantic background knowledge with two frames, \texttt{Movie} and
\texttt{Coop}(eration) (the latter 
to capture aspects like co-acting and co-directing).  
In addition, we trained the KALM-QA semantic
parser to recognize all of the 234,176 2- and 3-hop training questions
in MetaQA. This was done by selecting a small number of such questions,
training the parser on them (as in Section~\ref{structure_learning}), checking if there are still sentences
that KALM-QA cannot parse, selecting a few of those, train again, and so on.
Overall, it took a few rounds and required 88 sentences 
to generate 88 lvps that enabled KALM-QA to parse all of the training and
testing sentences in MetaQA (which is almost 300,000).

\paragraph{Results.}
KALM-QA achieved 100\% accuracy for both 2-hop and 3-hop questions---see Section
\ref{correctness}. 
This compares very favorably with the advanced machine learning approach that
introduced MetaQA,
VRN \cite{ZhangDKSS18}, which
reports the accuracy of 89.9\% for the 2-hop
dataset and 62.5\% for the 3-hop dataset. 
However, as detailed in Section~\ref{metaqa}, these results are
based on mislabeled data with high rate of incorrect query answers, so it is
possible that the real accuracy of this approach is 
lower
(it should be clear that KALM-QA only uses the ontology and the questions from MetaQA,
but not the mislabeled answers).

\paragraph{Discussion.}
The accuracy of 100\% achieved by KALM-QA may seem surprising not only
compared to machine learning approaches but also in light of our
previous work, which reported the accuracy of only about 95\%
\cite{Gao18,GaoFK18}. The explanation lies in the fact that the 5\% error
rate in out prior work was exclusively due to the noise and imprecision
present in BabelNet \cite{NavigliPonzetto:12aij}---the background ontology
used there.
In contrast, the present work uses the provided WikiMovies
ontology \cite{MillerFDKBW16}, which is precise and has no known errors
(not to be confused with the errors in MetaQA query answers).


\subsection{Correctness of KALM-QA Query Results}\label{correctness}

This subsection shows how we verified the correctness of the results produced by
KALM-QA for
such a large number of MetaQA queries---queries that required joining two
or three relations.
This was done by verifying that the actual Prolog queries produced by
KALM-QA are semantically correct and exactly match their corresponding
English sentences.

Since the number of such queries is large (close to 30,000), we had to
devise a method that exploits the symmetries present in MetaQA queries.
The method consists of three distinct phases:
\begin{enumerate}
\item  Identifying all the query templates that exist in KALM-QA
  translations (which are Prolog rules) of MetaQA
  queries (which are English sentences).
  \\
  It turned out that there are only 44 unique 2-hop query patterns and 59 unique 3-hop patterns.
\item Verifying that all MetaQA English questions that correspond to the
  same KALM-QA template differ only in the constant values used (like
  the named entities
  `\emph{Steven Spielberg}', `\emph{Bright Star}' or years like 1980).
  \\
  This means that these questions must be translated to logical queries
  that are instances of the same KALM-QA template.
\item For each KALM-QA query template $T$, let $MQA(T)$ be the set of
  all MetaQA questions that get translated into an instance of $T$.
  All that is needed now is to manually select one representative $q(T)\in
  MQA(T)$ for each $T$ and verify that its KALM-QA translation in Prolog
  is semantically correct.
  The number of such checks is a few hundreds and was accomplished manually
  within a few hours.
  \\
  Why hundreds and not just 44+59=103?
  Truth to be told, some templates had two to four different, semantically
  equivalent English forms in $MQA(T)$, so each of these forms had to be
  checked separately. For instance, 
  the sentences \emph{which films share the same actor of \textnormal{[}Bright Star\textnormal{]}}
  and \emph{what are the films that have the same actor of \textnormal{[}Friday the 13th\textnormal{]}}
  are semantically
  equivalent and thus have the same corresponding logical template.
\end{enumerate}
Additional details of the above steps are as follows.

\noindent
Step 1: A query like
\begin{verbatim}
    q(W2):-movie('FilmNm'=W2,'Id'=I2,'Actor'=W6),
           movie('FilmNm'='Bright Star','Id'=I8,'Actor'=W6),
           I2 \= I8.
\end{verbatim}
gets standardized into a template like
\begin{verbatim}
    q(A):-movie('FilmNm'=A,'Id'=B,'Actor'=C),
          movie('FilmNm'=xxxx,'Id'=D,'Actor'=C),
          B \= D.
\end{verbatim}
where the variables are standardized and the query-specific constants are
replaced with fixed constants. For example, \texttt{'Bright Star'} got replaced with
xxxx.
The resulting templates are then sorted lexicographically, which 
yielded the aforementioned number of templates.

\noindent
Step 2: MetaQA English sentences are then textually grouped according to
their corresponding KALM-QA templates. Each group is then scanned to verify
that all sentences there differ only in query-specific constant values.
This check is visual and rather tedious, but because the difference between
the different sentences in each group is trivial, the check takes only a
fraction of a second per line and took a few hours.

\noindent
Step 3: As mentioned, this required a manual check of a few hundred
MetaQA sentences with respect to their corresponding KALM-QA translations.

While KALM-QA yields 100\% correct answers,
we found that MetaQA misreports
a very large percentage of query answers, as discussed in Section \ref{metaqa}.
Most of this mislabeling is because MetaQA's designers failed to take into
account that the movie ontology has many distinct movies with the same
title, but there were also more subtle reasons.
An annotated report on all these errors appears in \url{https://github.com/tiantiangao7/kalm-qa}.
It should be noted that explanations for most of the errors in MetaQA were generated by an automated Prolog script.


\section{Related Work}\label{related_works}

We have already discussed the work on VRN \cite{ZhangDKSS18}, which used machine
learning to answer multi-hop questions provided by MetaQA. Other machine
learning approaches to query answering include the QA system of
\cite{BordesCW14} and KV-MemNN \cite{MillerFDKBW16}. In \cite{ZhangDKSS18},
it is reported than these two systems have much lower accuracy and thus,
presumably, than KALM-QA.

Non-machine-learning works include
the systems that are designed with a
multi-stage pipeline that translates NL questions into queries that use
SPARQL, SQL, or Prolog to then
query the actual databases. In these systems, translation is typically
driven by some kind of ontology.
Representative systems include ATHENA \cite{SahaFSMMO16},
PowerAqua \cite{LopezFMS12}, and NaLIR \cite{LiJ14}.
Compared to these works, KALM 
encompasses both knowledge authoring and question answering.
For the latter aspect, it would be interesting to compare the performance
of KALM with ATHENA and the others, and it
is also interesting to know how these systems would perform
on the MetaQA dataset.

The most obvious relation to KALM is, of course, Attempto itself
\cite{fuchs2008} and SBVR \cite{sbvr}, as they can also be used to query the Movie ontology.
However, since bare Attempto and SBVR include no background linguistic knowledge,
that knowledge would have to be supplied via background rules, e.g.,
\emph{if an actor appears in a film then the actor is an actor of the film}.
It is not hard to estimate (see \ref{bgrule_estimation})
that to capture the semantics expressed by the 88 lvps that KALM-QA had to
learn for MetaQA (Section~\ref{experiment}), Attempto and SBVR would require 974
background rules, hand-written in CNL (plus the rules for complex concepts
like cooperation, as in KALM-QA). A fruitful way to see the relationship
between KALM-QA and Attempto (and, to an extent, SBVR)
is to view the former as the
necessary semantic layer over the latter, which can make the combination
into a viable knowledge authoring and question answering technology.

SNOWY \cite{GomezHS94}, in development from mid 1980s to 2008, is related
to KALM-QA in that it does limited mediation between semantically
equivalent sentences with the help of an ontology, but likely still
requires many bridge rules to fully capture semantic equivalence.
Unfortunately, this system is no longer available and seems to have been
abandoned, so a more detailed comparison is hard to do.


\section{Conclusion}\label{conclusion}

This paper introduced KALM-QA, an extension of KALM that can answer complex
multi-hop CNL questions with very high accuracy.
Contrary to the prevailing trend, KALM-QA is based on logic programming
rather than machine learning.
Using MetaQA, a large collection of 2- and 3-hop questions against a large
movie ontology,
we demonstrated that KALM-QA achieves superior accuracy compared to
machine learning approaches. In fact, the error rate for the most accurate
machine
learning approach in our comparison turned out to be much higher
than what was reported in \cite{ZhangDKSS18} because the end-to-end learning
used in that work corrupted the expected query answers and the
system learned wrong answers for a very large subset of the questions
(north of 50\% for 3-hop queries).
For future work,
we plan to extend KALM with support for rule authoring 
and common sense and temporal reasoning.

\smallskip

\noindent
\textbf{Acknowledgments.}  
This work was partially supported by NSF grant 1814457.
We are grateful to Norbert Fuchs and Rolf Schwitter for enlightening discussions
on Attempto. Many thanks
to Katherine Choi for validating the bulk of
MetaQA queries.


\bibliographystyle{acmtrans}
\bibliography{main}

\newpage
\appendix

\section{Paraphrasing Questions with the Help of the Stanford Parser}\label{stanford_parser}

Although Attempto's grammar is fairly general and sufficient for most knowledge
authoring tasks, it is still too restrictive for question answering,
especially in the context of preexisting data sets like MetaQA
could be problematic in that many common and unambiguous NL
expressions are not accepted.
For instance, in the sentence
``\emph{Who appears in XYZ directed films},'' the
phrase \emph{XYZ directed films} would not be accepted by the ACE parser 
because the grammar requires that a noun can be modified only
by a preceding adjective or a relative clause that follows the noun.
Similarly, \emph{every film has actors} is not acceptable by that grammar
because it requires that every plural noun has to be preceded by a
determiner, like \emph{some}.
In this appendix, we show how the Stanford Parser can be used as a
paraphraser for ACE, making it possible to parse all of the MetaQA 260,000+
questions and also making the life easier for the KALM end user.

\begin{figure}[h]
 \centering
 \includegraphics[width=6cm]{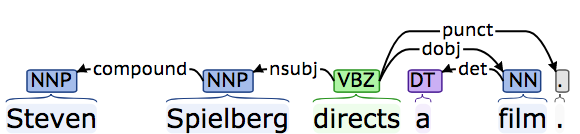}
 \caption{The dependency structure generated by Stanford Parser.}
 \label{figure:dependency_tree}
\end{figure}

The process works as follows.
First, the Stanford Parser is used to generate (1) the
\emph{part-of-speech} information for each word;
and (2) a set of dependency relations\footnote{\url{https://nlp.stanford.edu/software/dependencies_manual.pdf}}
that represent
the grammatical relations between pairs of words in the sentence.
Consider the sentence \emph{Steven Spielberg directs a film} whose
dependency structure is shown in Figure \ref{figure:dependency_tree}.
The dependency structure shows that there is a \emph{direct}-event (VBZ),
where the subject of the event (nsubj) is a compound noun (NNP) formed by the words 
\emph{Steven} and \emph{Spielberg}, and the object of the event (dobj)
is \emph{film}.
Based on these dependency relations, KALM uses a set of transformation
rules to recognize a subset of the dependency relations 
whose corresponding English features
are not supported by ACE parser and then paraphrase them into the
corresponding ACE phrases.
Examples of the transformation rules are shown below:
Understandably, these particular rules were driven by the features in
MetaQA that are not supported by ACE, but additional rules can be added, if
desired. 
\begin{itemize}
\item \emph{Identifying past participle clauses and paraphrasing them as relative clauses}.
Consider the sentence ``\emph{Who watched a film directed by Steven
  Spielberg?}'' 
The phrase \emph{directed by Steven Spielberg} is a past participle clause
that modifies the nominal \emph{film}.
KALM identifies the dependency relations that correspond to the past
participle clauses
in the sentence and then paraphrases them as relative clauses
by adding the phrase \emph{that is/are} after the modified nominal.

\item \emph{Accepting verbs of different tenses}. ACE accepts verbs in the
  present tense only, which is a serious inconvenience in practice not to
  mention that MetaQA uses past tense almost everywhere.
  KALM recognizes the tense information of verbs via the
  \emph{part-of-speech} tags produced by the Stanford Parser
  and then normalizes all verbs into the present tense. 
  Since currently KALM does not support temporal reasoning, this change of
  verb tense does not affect  question answering.

\item \emph{Adding missing articles}. ACE insists that a singular countable
  noun must be preceded by an article (\emph{a}, \emph{an},
  \emph{the}) unless it is modified by possessive determiners, like
  \emph{John's}, or quantifiers, like \emph{one}. The Stanford Parser uses the
  dependency relations \emph{det}, \emph{nmod:poss}, and \emph{nummod} to
  represent the articles, possessive determiners, and quantifiers
  receptively.  If none of these relations is found, KALM adds the article
  \emph{a}/\emph{an} before the noun.

\item \emph{Recognizing compound proper nouns}.  The ACE parser cannot
  recognize compound proper nouns (e.g., \emph{New York}, \emph{Steven
    Spielberg}) unless the user manually inserts a hyphen (\emph{-})
  between the components (e.g., \emph{New-York}, \emph{Steven-Spielberg}).
  KALM identifies the compound proper nouns by finding the \emph{compound}
  relations from the dependency parses, as in
  Figure~\ref{figure:dependency_tree}, and then creates compound proper
  nouns acceptable to ACE.
\end{itemize}

\noindent
In addition to the above transformation rules, there are also a few ad-hoc,
MetaQA-specific
rules to handle edge cases with which the Stanford parser could not help.
Further details can be found in \url{https://github.com/tiantiangao7/kalm-qa}.


\newpage
\section{CNL's Background Rules vs. KALM's LVPs}
\label{bgrule_estimation}

This appendix provides an estimate for the number of handwritten rules that
would have to be supplied in order to be able to answer MetaQA queries
without KALM-QA. The estimate is then used to argue that KALM-QA is a
much more practical alternative to writing background linguistic knowledge
in CNL.

Tables \ref{stat:movie_frame} and \ref{stat:coop_frame} show the numbers of
lvps that correspond to each pair of roles in the frames
\texttt{Movie} and \texttt{Coop}.
Note that, to represent semantic equivalence, every role-role-lvp triple in
these tables requires two background rules in plain CNL.
For instance, in the \texttt{Movie} frame, the triple
\texttt{Film}-\texttt{Actor}-one-of-the-lvps would require
both of these bridging background rules:
(1) \emph{if an actor appears in a film then the actor is an actor of the
  film}; and
(2) \emph{if an actor is an actor of a film then the actor appears in the film}.

In total, both frames (that are represented by 88 lvps obtained from 88 annotated
sentences) do the work of 974 background rules like the above.
The number of required background rules will further increase, if we consider the need to represent
information embedded in KALM's built-in frames like \texttt{Distinct},
which capture the concept of distinctness of entities, etc.

Thus, the amount of information needed to capture the domain of movies
using KALM-QA is less than 10\% of what is needed using plain CNL, not
armed with such a semantic layer. Moreover, we posit that the amount of
manual work to build one lvp is significantly less than the work required
to write a background rule in CNL, especially if tooling support is
developed for KALM-QA.
\begin{table}[h!]
\centering
\begin{tabular}{ccc}
\hline \hline
Role1 & Role2 & Number of lvps\\ \hline
Film & Actor & 17 \\
Film & Writer & 21 \\
Film & Director & 14 \\
Film & Release Year & 6 \\
Film & Genre & 7 \\
Film & Language & 4 \\ \hline\hline
\end{tabular}
\caption{\label{stat:movie_frame}\textnormal{Statistics for the \texttt{Movie} Frame}}
\end{table}

\begin{table}[h!]
\centering
\begin{tabular}{ccc}
\hline \hline
Role1 & Role2 & Number of lvps\\ \hline
Actor & Actor & 4 \\
Writer & Writer & 1 \\
Director & Director & 1 \\
Actor & Writer & 2 \\
Actor & Director & 2 \\
\hline\hline
\end{tabular}
\caption{\label{stat:coop_frame}\textnormal{Statistics for the \texttt{Coop} Frame}}
\end{table}


\newpage
\label{lastpage}
\end{document}